\documentclass[11pt]{article}

\usepackage[preprint]{acl}

\usepackage{times}
\usepackage{latexsym}

\usepackage[T1]{fontenc}

\usepackage[utf8]{inputenc}

\usepackage{microtype}

\usepackage{inconsolata}

\usepackage{graphicx}

\usepackage{booktabs}
\usepackage{linguex}
\usepackage{enumitem}
\usepackage{xcolor}
\usepackage[export]{adjustbox}
\usepackage{caption}
\usepackage{subcaption}
\usepackage{placeins}
\usepackage{afterpage}
\usepackage{tikz}

\definecolor{myteletypecolor}{RGB}{0, 102, 204} 
\newcommand{\query}[1]{{\ttfamily\color{myteletypecolor}#1}}

\newcommand{\filter}[1]{\textcolor{red}{\textsc{#1}}}

\newcommand{\toggle}[1]{\textcolor{darkblue}{\textsc{#1}}}

\title{DiscoExplorer: \\An Open Interface for the Study of Multilingual Discourse Relations}

\author{Amir Zeldes \\
  Georgetown University \\
  \texttt{amir.zeldes@georgetown.edu} \\}

\begin{document}
\maketitle
\begin{abstract}
The relations connecting propositions in discourse such as \textsc{cause} (A because B) or \textsc{concession} (A although B) are a subject of intense interest in Computational Linguistics and Pragmatics, but challenging to study and compare across languages. Recent progress in standardizing discourse relation inventories across datasets offers the potential to facilitate such studies, but is hindered by the complexity of relevant data and the lack of easily accessible interfaces to analyze it. In this paper we present DiscoExplorer, a new open source web interface, capable of running on local computers,  which we use to make datasets from the DISRPT Shared Task on discourse relation classification publicly available, covering 16 different languages. We present the query language, search and visualization facilities for relations and signaling devices such as connectives, as well as some example studies.
\end{abstract}

\section{Introduction}

Discourse relations are the implicit and explicit semantic/pragmatic connections that arise when multiple propositions are juxtaposed in a text or conversation. For example, in \ref{ex:explicit-implicit}, the explicit connective `when' indicates a \textsc{temporal} relation between the two arguments 1 and 2, while the \textsc{causal} relation between 1 and 3 is understood implicitly (Jin is upset \textit{because} Kim left).

\ex. $[$\textit{Kim left}$]_1$ $[$\textit{when Jin arrived.}$]_2$ $[$\textit{Jin is upset now.}$]_3$ \label{ex:explicit-implicit}

A variety of theories have attempted to describe discourse relations and construct datasets for their study, including Rhetorical Structure Theory (RST, \citealt{MannThompson1988}), Segmented Discourse Representation Theory (SDRT, \citealt{AsherLascarides2003}), the Penn Discourse Treebank (PDTB, \citealt{prasad-etal-2014-reflections}) and discourse dependencies \cite{morey-etal-2018-dependency}. However because each theory and dataset has tended to use distinct relation inventories and data structures (for example hierarchical trees, graphs, or pairs of text spans), comparisons across languages or even datasets in the same language have been challenging.

More recently, the DISRPT shared task \cite{braud-etal-2024-disrpt} has made progress in unifying data from such formalisms by focusing on what they have in common: the postulation of relations between parts of a text, and optional inclusion of information about signaling devices, for example the distinction between implicit and explicit relations above. In its most recent edition the shared task also unified relation labels across 38 datasets in 16 different languages \cite{braud-etal-2025-disrpt}, facilitating cross-linguistic comparisons for the first time, similarly to initiatives to consolidate labels for describing multilingual syntactic functions in projects such as Universal Dependencies (UD, \citealt{de-marneffe-etal-2021-universal}). However what has been missing compared to projects like UD is an easily accessible interface to search and compare data, identify errors, and visualize patterns in datasets. The main contributions of this short paper aim to fill this gap:

\begin{itemize}[itemsep=0.8pt]
    \item We provide a high performance, open source, client-side interface in pure JavaScript that can be run on any PC
    \item We make the datasets from the DISRPT shared task searchable online for the public
    \item We propose a simple, flexible query language to facilitate access for new users
\end{itemize}

\section{Related work}

While many local graph search tools exist for linguistic data, such as Semgrex, Ssurgeon or Semgrex-Plus \cite{tamburini-2017-semgrex,bauer-etal-2023-semgrex}, almost all are limited to searching within sentence boundaries, and are therefore not capable of representing relations across entire texts. Several online interfaces have facilitated search in syntactically and even semantically annotated treebanks \cite{guibon-etal-2020-collaborative,amblard-etal-2022-graph}, but dedicated interfaces for discourse relations are rare, and have generally been fitted to a single resource and theory, such as the Spanish \cite{DBLP:conf/ranlp/CunhaMSCRB11} and Basque \cite{iruskieta2013rstbasque}  RST treebank interfaces. Converters for RST data exist to enable searching through data using ANNIS \cite{KrauseZeldes2016}, a generic multilayer corpus search interface. However the system is considerably heavier, slower and has a complex query language which is not tailored to discourse relations, and currently cannot import data from other discourse formalisms or the DISRPT format.

Our work takes its primary inspiration from the Grew Match search interface for UD treebanks \cite{guibon-etal-2020-collaborative}, which leverages the consistent format and label inventory of the UD project to allow access to treebanks using a consistent query language and architecture.

\section{DiscoExplorer}

\subsection{Architecture}

Our architecture is designed with three goals in mind: 1. minimizing compute costs to prevent needing a dedicated (and expensive) server; 2. making it possible to run the interface locally for users with proprietary data that cannot be exposed online; and 3. running a fast and responsive search with minimal dependencies. To achieve these goals, we implemented a client-side solution in JavaScript using React, without a database backend, no dedicated indexing (e.g.~Meilisearch) or visualization libraries (e.g.~D3.js). Instead, we focus on using pure JavaScript, HTML and CSS wherever possible to ensure stability and longevity of the software.

Our data model focuses on discourse relations as the instance to be searched over, where relations are aligned to token positions in documents and span over two possibly discontinuous argument spans (e.g.~the cause and effect for \textsc{causal} relations). Relations that do not cover entire sentences are also associated with context spans indicating words before, after or between the arguments within the same sentences, ensuring that full sentence context is provided with each match. Finally, relations carry labels, a direction (1>2 or 1<2) and possibly a list of typed and subtyped signal tokens, for datasets marking connectives or other signal types.

\subsection{Basic interface}

\begin{figure*}[t!bh]
\centering
\includegraphics[width=0.9\textwidth, frame]{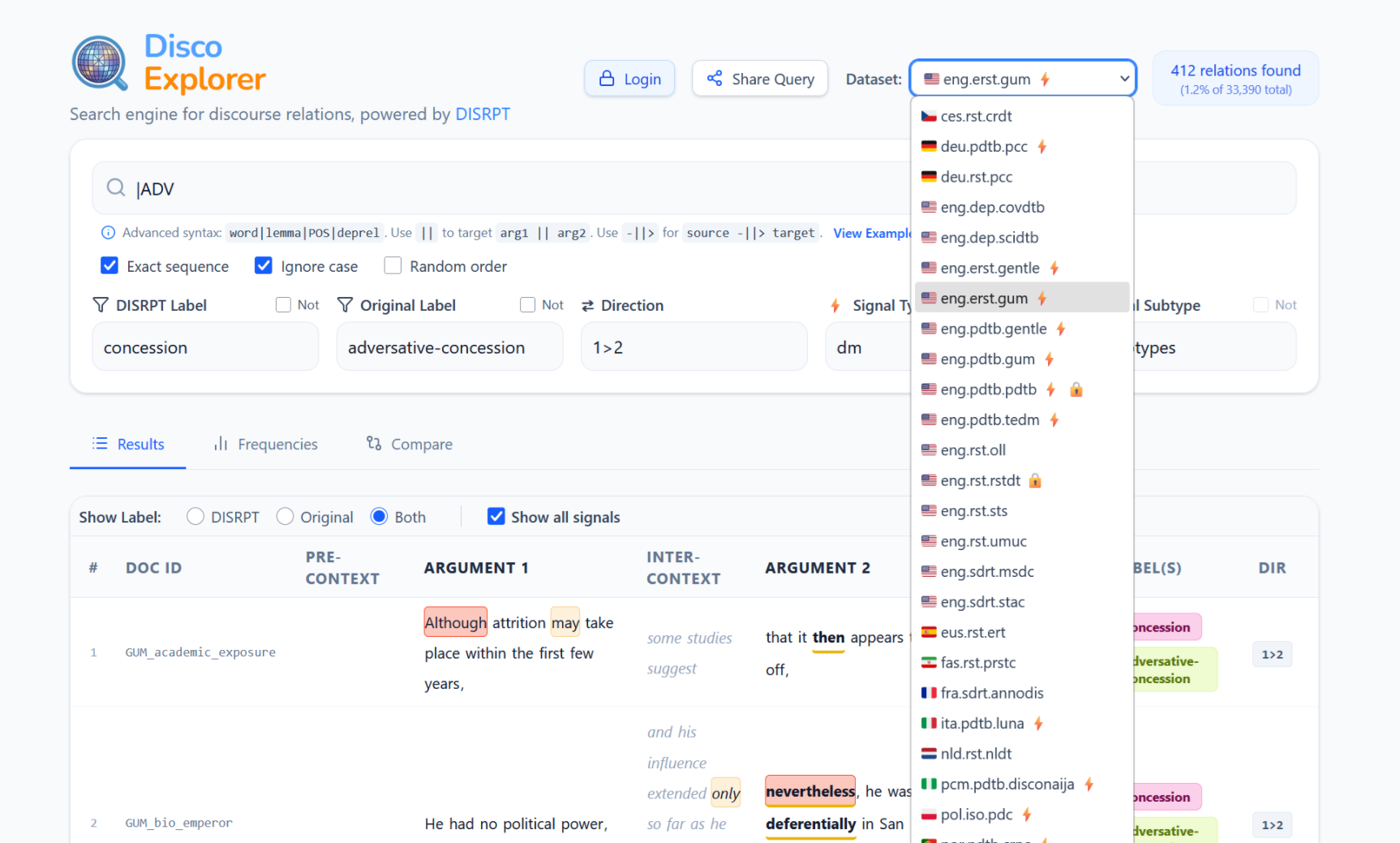}
\vspace{-4pt}
\caption{DiscoExplorer search interface: Users can input a query and select filters. Underlines show query matches and signals are highlighted (e.g.~red for discourse markers, yellow for lexical signals).}
\label{fig:search-interface}
\end{figure*}

The web interface is arranged around two areas: the query form at the top of Figure \ref{fig:search-interface}, and the results area at the bottom, which can display concordances for qualitative searches, or switch to a `frequencies' tab for quantitative analysis. The interface was initially tested with students in a seminar on computational models of discourse at Georgetown University (LING-8415), and based on student feedback, an additional tab was added to perform comparisons between datasets. We are also planning to collect feedback from CODI attendees and the DISRPT community to develop additional features.

The basic query form is meant to be user friendly by exposing the available datasets and labels in each dataset as drop down filters. Negation of a filter is realized as a simple checkbox -- for example, selecting the label \textsc{condition} and the negation box for any signal type in \texttt{eng.erst.gum} (data from the GUM corpus, \citealt{zeldes2017gum}) yields examples of implicit conditionals, as in \ref{ex:implicit-condition}. More complex queries must utilize the DiscoExplorer query language (DEQL) described in \ref{sec:ql}.

\ex. $[$\textit{you take this painting,}$]_1$ $[$\textit{I want that recorder}$]_2$ (=if you take this painting) \label{ex:implicit-condition}

\subsection{Query language -- DEQL}\label{sec:ql}

Our query language aims to be simple but powerful, meaning on the one hand, it should respond as expected to simply typing words in the search box, while on the other hand allowing users to do exact sequence or flexible match queries, queries restricted to the first/second or source/target spans of the relation, as well as leveraging token annotations. Since DISRPT data is released with accompanying UD annotations, we expose the UD POS tags, dependency labels and lemmas directly for querying. All queries can be restricted by the UI to a specific relation label chosen from a drop down list (either the universal DISRPT label, or each dataset's original labels, or both), specific signal types or subtypes if available in the data (e.g.~explicit connectives), and relation directions. The exact query can be saved and reproduced via a shareable link.

As an example of simple text based queries and their interactions with argument spans, consider the differences between the following, all executed with the `exact sequence' match turned off and the \textsc{condition} label selected:\footnote{UI filters are indicated in red and are not part of the query string, but are stored in reproducible shareable query links.}

\ex. \filter{condition} \query{if then} (finds \textsc{condition} relations with `if' and `then' anywhere)\label{q:if-then-anywhere}

\ex. \filter{condition} \query{if || then} (same, but ensures `if' and `then' are in arg1 and arg2)\label{q:if-then-ordered}

\ex. \filter{condition} \query{if -||> then} (same, but `if' must be in the relation source and `then' in the target, regardless of text order)\label{q:if-then-src}

While in \ref{q:if-then-anywhere} we only guarantee that `if' and `then' appear somewhere, in \ref{q:if-then-ordered} we require that they appear in that text order, one in each argument. By contrast, \ref{q:if-then-src} requires that `if' appears in the source of the relation (the protasis) and `then' in the target (the apodosis), regardless of their text order.

More experienced users who are familiar with UD annotations may also want to use token annotations to restrict queries. To enable this we use the format \query{word|lemma|pos|deprel}, where each of these elements may be lacking. If less than three annotations are specified, the system uses the search values to implicitly identify the key, since POS and deprel have closed vocabularies. Thus the following searches find:

\ex. \filter{purpose} \toggle{exact} \query{to|PART |VERB|advcl -||>} (\textsc{purpose} relation with a to-infinitive)\label{q:inf}

\ex. \filter{temporal} \toggle{exact} \query{when |ADJ|advcl -||>} (\textsc{temporal} with `when' followed by a reduced adjectival adverbial clause)\label{q:when}

The example in \ref{q:inf} will find VERBs heading an adverbial clause (UD \textit{advcl}) immediately preceded by the word `to' tagged as PART. The interface automatically detects that VERB is a POS tag value and \textit{advcl} is a dependency relation. In \ref{q:when} we find reduced temporal clauses of the type `when possible', since the word `when' must be followed immediately by an adjective heading an adverbial clause. The final operator \query{-||>} ensures that both searches only consider the source span of the relation, regardless of text order.

\subsection{Frequencies interface}

\begin{figure}[h!tb]
\centering
\includegraphics[width=\columnwidth, frame]{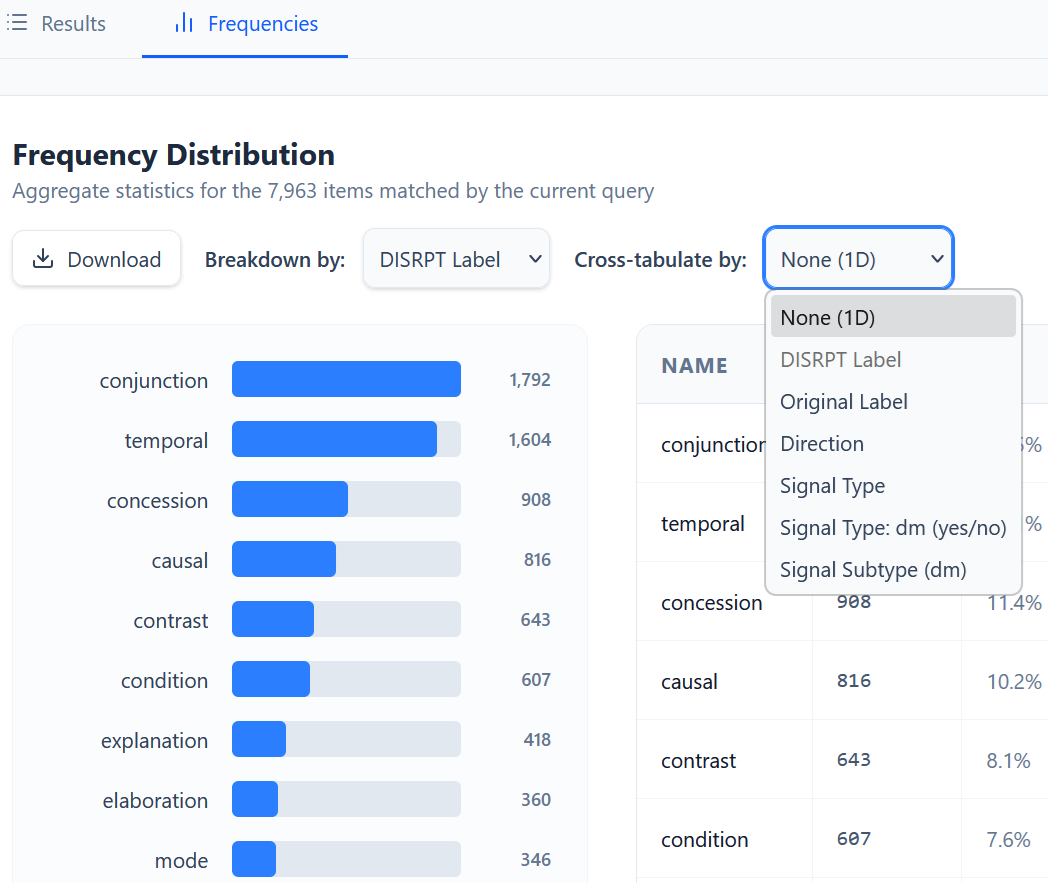}
\caption{Frequency breakdown of DISRPT labels.}
\label{fig:freqs}
\vspace{-8pt}
\end{figure}

The frequencies tab gives raw counts, percentages and plots of a category or numerical variable selected by the user from the Breakdown drop down (see Figure \ref{fig:freqs}). Categorical variables include DISRPT labels, original labels, relation direction, signal type/subtype and any available metadata (for example genre, if known). If filters are selected for any of these in the query, a binary yes/no breakdown of the selected feature is also available. Updating the query instantly updates matches, numbers and plots, and raw results are also downloadable as a .tsv file.

\begin{figure}[h!tb]
\centering
\includegraphics[width=\columnwidth, frame]{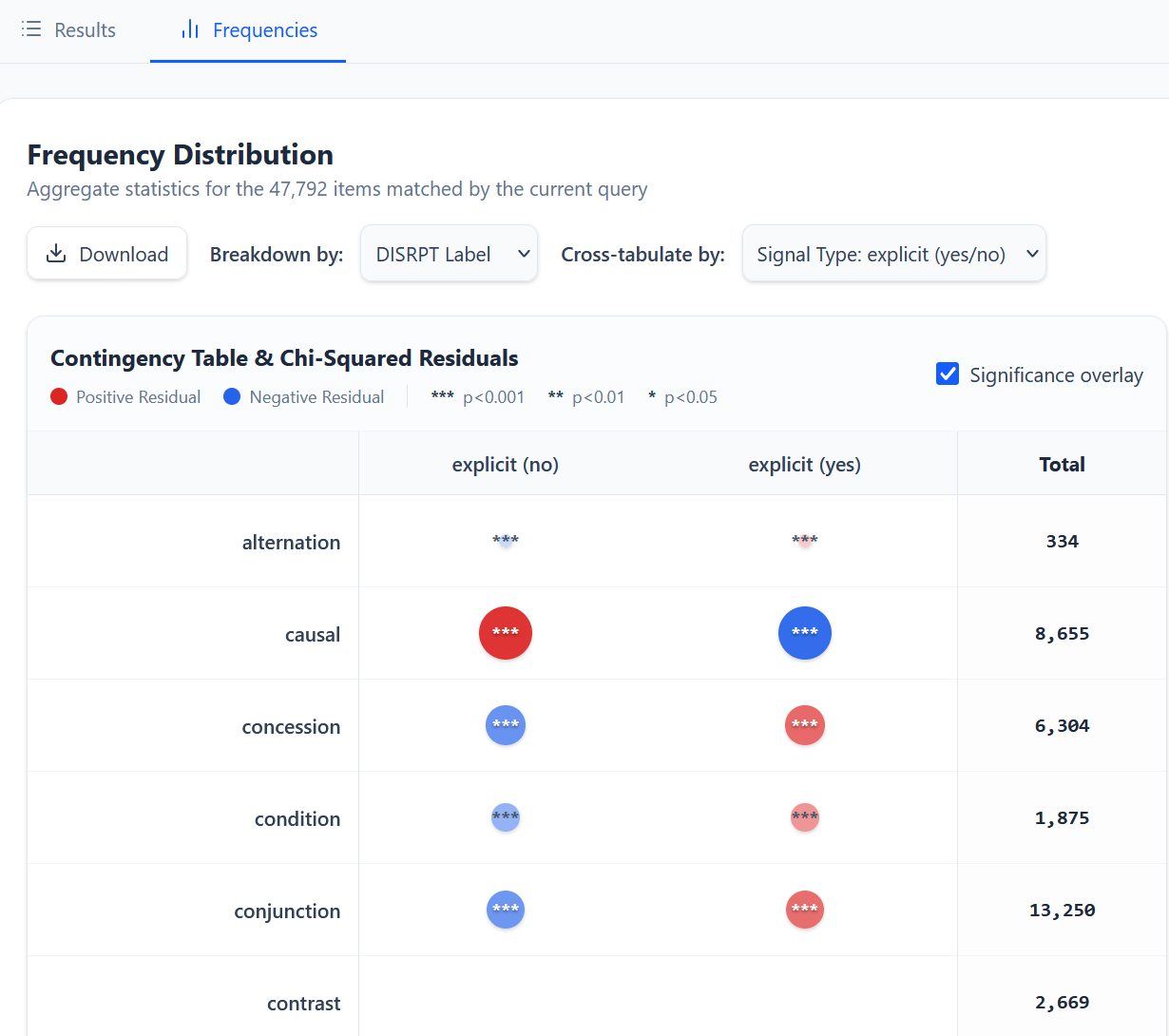}
\caption{Association of explicitness vs. label in PDTB.}
\label{fig:crosstab}
\end{figure}

A second drop down called `Cross-tabulate' allows users to select a second dimension from the same options and generate a cross table, coupled with a chi-squared residual plot indicating combinations that appear more or less than expected, as well as displaying significance codes. For example, Figure \ref{fig:crosstab} shows an association plot of explicit connective signals vs. DISRPT label in the English PDTB corpus, showing that while \textsc{concession}, \textsc{condition} and \textsc{conjunction} are mostly explicit, \textsc{causal} relations are more often implicit, while \textsc{contrast} relations are more balanced.

If a numerical variable is chosen for breakdown, the interface will plot a boxplot (for a single variable) or scatterplot (two cross-tabulated numerical variables) or multiple boxplots (numerical cross-tabulated with categorical). Available numerical variables are currently argument length in tokens and percentile position in document (for argument 1 or 2 in text order), the same for the source or target argument (regardless of text order), distance in tokens between arguments, and the number of signals for the relation (if available).

\subsection{Comparison interface}

\begin{figure}[h!tb]
\centering
\includegraphics[width=\columnwidth, frame]{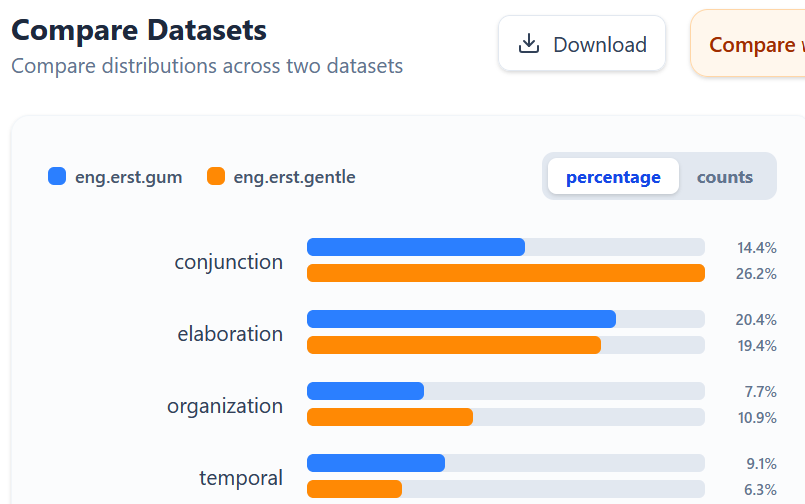}
\caption{Relation labels in GUM vs. GENTLE.}
\label{fig:comparison}
\end{figure}

Based on student feedback, comparing datasets is a desirable capability, and we implement this in a similar way to cross-tabulation, where, instead of using a categorical variable, we use dataset identity. However, since each dataset has its own distribution for each variable, we display results for each value side-by-side, with the primary selected dataset in blue and the comparison in orange with pairwise plots, as shown for a categorical variable (label type) with barplots in Figure \ref{fig:comparison} for a comparison between the eRST GUM corpus and the eRST GENTLE corpus (Genre Tests for Linguistic Evaluation, \citealt{aoyama-etal-2023-gentle}), which follows the same annotation scheme but includes 8 challenging genres such as medical texts, poetry and even course syllabuses.

The figure shows that \textsc{conjunction} is more common in  GENTLE (in orange), which is primarily due to genres containing many lists, such as medical notes and syllabuses. The \textsc{elaboration} label, but contrast, is very similar in prevalence.

\begin{figure}[h!tb]
\centering
\includegraphics[width=0.85\columnwidth, frame]{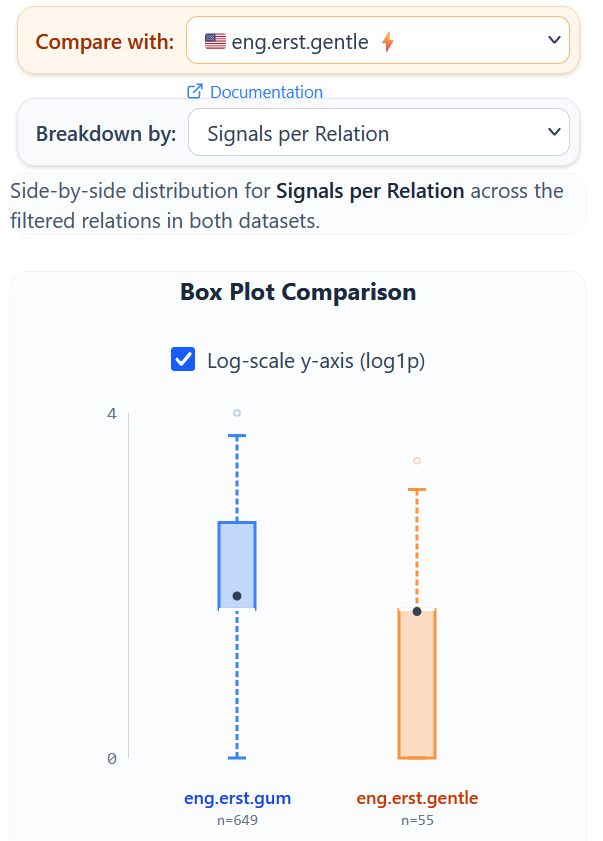}
\caption{Signals per \textsc{mode} relation compared.}
\label{fig:comparison2}
\vspace{-10pt}
\end{figure}

As with frequencies, numerical variables receive side-by-side boxplots. Figure \ref{fig:comparison2} shows the number of signals per relation, this time filtered to show just \textsc{mode} relations (manner and means). These have significantly fewer signals in GENTLE, largely owing to data from the poetry and medical genres.

\section{Evaluation}

\paragraph{Data} We import the 38/39 DISRPT 2025 datasets which contain discourse relations (the remaining dataset contains only discourse unit segmentation information, without labels). The datasets come from five different frameworks: RST, PDTB, SDRT, eRST \cite{zeldes-etal-2025-erst} and discourse dependencies. In total, these cover over 300,000 relations across 5 million tokens in almost 10,000 documents (see Table \ref{tab:datasets} in Appendix \ref{sec:appendix-datasets} for full details). The largest dataset is English PDTB \cite{prasad-etal-2014-reflections} with over 47K relations and 1.1M tokens, as well as five signal types (explicit or implicit connectives, alternative lexicalizations and constructions, and a special hypohphora type for questions). Datasets in the eRST framework also distinguish signal subtypes, in a taxonomy of 8 major types and over 40 subtypes.

\paragraph{Performance} While it is difficult to benchmark our system due to a lack of directly comparable alternatives, we conduct a simple timing experiment using the GUM data in comparison to its publicly available version in ANNIS. We note that ANNIS can perform much more elaborate searches than DiscoExplorer, such as dependency graph queries between tokens (e.g.~checking that a token is the subject of a specific verb), as well as querying other annotation layers, such as entity annotations; here we limit comparison to simple searches for tokens and discourse relations on a consumer laptop. Since our data is loaded into main memory, query response times are close to instantaneous, with the only added latency of loading the dataset once, which ANNIS does not have (Table \ref{tab:performance}).

\begin{table}[h!tb]
\resizebox{\columnwidth}{!}{
\begin{tabular}{llrrr}
\toprule
\textbf{Query   type} & \textbf{DEQL}              & \textbf{DiscoExp.} & \textbf{ANNIS} & \textbf{Hits} \\
\midrule
(load)                & --                         & 2.820s                  & --             & --               \\
tok                   & \query{think}                      & 0.022s                 & 3.98s          & 291              \\
tok+pos+deprel        & \query{think|VERB|advcl}           & 0.027s                 & 4.68s          & 17               \\
rel+tok+pos           & \filter{CONJUNCTION} \query{think|VERB}     & 0.030s                 & 3.01s          & 12               \\
neg-rel+tok+pos       & \toggle{NOT} \filter{CONJUNCTION} \query{think|VERB} & 0.028s                 & 4.41s          & 410              \\
rel                   & \filter{ELABORATION}                & 0.003s                 & 3.45s          & 6812            \\
\bottomrule
\end{tabular}
}
\caption{Query latency compared with ANNIS.}
\label{tab:performance}
\end{table}

\section{Discussion and Conclusion}

This short paper presented DiscoExplorer, a new browser based search interface for multilingual discourse relation datasets based on the DISRPT benchmark. The interface offers a simple user-friendly way to search for discourse relations and examine their distributions using filters, as well as a more complex query language to restrict matches by tokens and their annotations. Quantitative results can be tabulated, plotted and downloaded.

A comparison of query run times with ANNIS showed that although the interface requires an initial load time for each dataset, the approach using main memory search in JavaScript is very fast. While some of the complex searches a system like ANNIS would allow are not supported, tailoring the interface to the DISRPT data model, centered around discourse relation instances, allows for a simple query language and data structure, and also means we do not need a backend or any compute resources to offer the system to the public. 

A further advantage of relying on the DISRPT data model is the abundance of data already available in the shared task format (currently 38 datasets), and the likely release of further public data in the format of the shared task, which has been running for four iterations as of 2026. 

As a result of submissions to the task, and especially the transition to multilingual models trained on all datasets, it is also increasingly possible to generate predicted datasets following the DISRPT label scheme in a variety of languages. The current best performing system for predicting relation labels, DeDisCo \cite{ju-etal-2025-dedisco}, achieves 76.13\% accuracy on the DISRPT test set across languages, and 75.55\% on Chinese RST \cite{peng-etal-2022-gcdt}, 79.17\% on Portuguese \cite{CRPC-DB-Portuguese}, or 71.39\% on the PDTB framework Georgetown Discourse Treebank (GDTB, \citealt{liu-etal-2024-gdtb}), suggesting that automatically tagged corpora that are useful for research on discourse relations across languages may not be far and could easily be made searchable using this system. 

We release the code and place the system online for public use together with this paper, available at \url{https://gucorpling.org/discoexplorer}.

\section*{Limitations}

This paper only presents and evaluates a search interface in terms of responsiveness and does not conduct a user study, though we hope to gather some feedback on the system from participants at the CODI workshop and the community using DISRPT data. The data used by the system is provided by DISRPT as-is, and we make no claims regarding the accuracy of particular annotations within those datasets. AI was not used in any way to write this paper, though AI coding assistants were used in the creation and debugging of the system itself.


\bibliography{custom}

@Article{MannThompson1988,
  title                    = {Rhetorical {S}tructure {T}heory: Toward a Functional Theory of Text Organization},
  author                   = {William C. Mann and Sandra A. Thompson},
  journal                  = {Text},
  year                     = {1988},
  number                   = {3},
  pages                    = {243--281},
  volume                   = {8}
}

@Book{AsherLascarides2003,
  author    = {Nicholas Asher and Alex Lascarides},
  publisher = {Cambridge University Press},
  title     = {Logics of Conversation},
  year      = {2003},
  address   = {Cambridge},
  series    = {Studies in Natural Language Processing}
}

@article{prasad-etal-2014-reflections,
    title = "Reflections on the {P}enn {D}iscourse {T}ree{B}ank, Comparable Corpora, and Complementary Annotation",
    author = "Prasad, Rashmi  and
      Webber, Bonnie  and
      Joshi, Aravind",
    journal = "Computational Linguistics",
    volume = "40",
    number = "4",
    month = dec,
    year = "2014",
    address = "Cambridge, MA",
    publisher = "MIT Press",
    url = "https://aclanthology.org/J14-4007",
    doi = "10.1162/COLI_a_00204",
    pages = "921-950",
}

@article{morey-etal-2018-dependency,
    title = "A Dependency Perspective on {RST} Discourse Parsing and Evaluation",
    author = "Morey, Mathieu  and
      Muller, Philippe  and
      Asher, Nicholas",
    journal = "Computational Linguistics",
    volume = "44",
    number = "2",
    month = jun,
    year = "2018",
    address = "Cambridge, MA",
    publisher = "MIT Press",
    url = "https://aclanthology.org/J18-2001",
    doi = "10.1162/COLI_a_00314",
    pages = "197-235",
}

@inproceedings{guibon-etal-2020-collaborative,
    title = "When Collaborative Treebank Curation Meets Graph Grammars",
    author = {Guibon, Ga{\"e}l  and
      Courtin, Marine  and
      Gerdes, Kim  and
      Guillaume, Bruno},
    editor = "Calzolari, Nicoletta  and
      B{\'e}chet, Fr{\'e}d{\'e}ric  and
      Blache, Philippe  and
      Choukri, Khalid  and
      Cieri, Christopher  and
      Declerck, Thierry  and
      Goggi, Sara  and
      Isahara, Hitoshi  and
      Maegaard, Bente  and
      Mariani, Joseph  and
      Mazo, H{\'e}l{\`e}ne  and
      Moreno, Asuncion  and
      Odijk, Jan  and
      Piperidis, Stelios",
    booktitle = "Proceedings of the Twelfth Language Resources and Evaluation Conference",
    month = may,
    year = "2020",
    address = "Marseille, France",
    publisher = "European Language Resources Association",
    url = "https://aclanthology.org/2020.lrec-1.651/",
    pages = "5291--5300",
    language = "eng",
    ISBN = "979-10-95546-34-4"
}

@inproceedings{amblard-etal-2022-graph,
    title = "Graph Querying for Semantic Annotations",
    author = "Amblard, Maxime  and
      Guillaume, Bruno  and
      Pavlova, Siyana  and
      Perrier, Guy",
    editor = "Bunt, Harry",
    booktitle = "Proceedings of the 18th Joint ACL - ISO Workshop on Interoperable Semantic Annotation within LREC2022",
    month = jun,
    year = "2022",
    address = "Marseille, France",
    publisher = "European Language Resources Association",
    url = "https://aclanthology.org/2022.isa-1.13/",
    pages = "95--101"
}

@inproceedings{braud-etal-2024-disrpt,
    title = "{DISRPT}: A Multilingual, Multi-domain, Cross-framework Benchmark for Discourse Processing",
    author = "Braud, Chlo{\'e}  and
      Zeldes, Amir  and
      Rivi{\`e}re, Laura  and
      Liu, Yang Janet  and
      Muller, Philippe  and
      Sileo, Damien  and
      Aoyama, Tatsuya",
    editor = "Calzolari, Nicoletta  and
      Kan, Min-Yen  and
      Hoste, Veronique  and
      Lenci, Alessandro  and
      Sakti, Sakriani  and
      Xue, Nianwen",
    booktitle = "Proceedings of the 2024 Joint International Conference on Computational Linguistics, Language Resources and Evaluation (LREC-COLING 2024)",
    month = may,
    year = "2024",
    address = "Torino, Italia",
    publisher = "ELRA and ICCL",
    url = "https://aclanthology.org/2024.lrec-main.447/",
    pages = "4990--5005"
}

@inproceedings{braud-etal-2025-disrpt,
    title = "The {DISRPT} 2025 Shared Task on Elementary Discourse Unit Segmentation, Connective Detection, and Relation Classification",
    author = "Braud, Chlo{\'e}  and
      Zeldes, Amir  and
      Li, Chuyuan  and
      Liu, Yang Janet  and
      Muller, Philippe",
    editor = "Braud, Chlo{\'e}  and
      Liu, Yang Janet  and
      Muller, Philippe  and
      Zeldes, Amir  and
      Li, Chuyuan",
    booktitle = "Proceedings of the 4th Shared Task on Discourse Relation Parsing and Treebanking (DISRPT 2025)",
    month = nov,
    year = "2025",
    address = "Suzhou, China",
    publisher = "Association for Computational Linguistics",
    url = "https://aclanthology.org/2025.disrpt-1.1/",
    doi = "10.18653/v1/2025.disrpt-1.1",
    pages = "1--20",
    ISBN = "979-8-89176-344-9"
}

@article{de-marneffe-etal-2021-universal,
    title = "{U}niversal {D}ependencies",
    author = "de Marneffe, Marie-Catherine  and
      Manning, Christopher D.  and
      Nivre, Joakim  and
      Zeman, Daniel",
    journal = "Computational Linguistics",
    volume = "47",
    number = "2",
    month = jun,
    year = "2021",
    address = "Cambridge, MA",
    publisher = "MIT Press",
    url = "https://aclanthology.org/2021.cl-2.11/",
    doi = "10.1162/coli_a_00402",
    pages = "255--308"
}

@inproceedings{DBLP:conf/ranlp/CunhaMSCRB11,
  author       = {Iria da Cunha and
                  Juan{-}Manuel Torres{-}Moreno and
                  Gerardo Sierra and
                  Luis Adri{\'{a}}n Cabrera{-}Diego and
                  Brenda Gabriela Castro Rol{\'{o}}n and
                  Juan Miguel Rolland Bartilotti},
  editor       = {Galia Angelova and
                  Kalina Bontcheva and
                  Ruslan Mitkov and
                  Nicolas Nicolov},
  title        = {The {RST} Spanish Treebank On-line Interface},
  booktitle    = {Recent Advances in Natural Language Processing, {RANLP} 2011, 12-14
                  September, 2011, Hissar, Bulgaria},
  pages        = {698--703},
  publisher    = {{RANLP} 2011 Organising Committee},
  year         = {2011},
  url          = {https://aclanthology.org/R11-1101/},
  bibsource    = {dblp computer science bibliography, https://dblp.org}
}

@inproceedings{iruskieta2013rstbasque,
  author = {Iruskieta, Mikel and
            Aranzabe, Mar{\'i}a Jes{\'u}s and
            Diaz de Ilarraza, Arantza and
            Gonzalez-Dios, Itziar and
            Lersundi, Mikel and
            Lopez de Lacalle, Oier},
  title = {The {RST} {Basque TreeBank}: {An} Online Search Interface to Check Rhetorical Relations},
  booktitle = {Anais do IV Workshop ``A RST e os Estudos do Texto''},
  pages = {40--49},
  address = {Fortaleza, CE, Brasil},
  month = oct,
  year = {2013},
  note = {21--23 October 2013},
  publisher = {Sociedade Brasileira de Computa{\c{c}}{\~a}o}
}

@Article{KrauseZeldes2016,
  author    = {Thomas Krause and Amir Zeldes},
  title     = {{ANNIS3}: {A} New Architecture for Generic Corpus Query and Visualization},
  journal   = {Digital Scholarship in the Humanities},
  year      = {2016},
  volume    = {31},
  number    = {1},
  pages     = {118--139}
}

@article{zeldes2017gum,
  title={The {GUM} corpus: {C}reating multilayer resources in the classroom},
  author={Zeldes, Amir},
  journal={Language Resources and Evaluation},
  volume={51},
  number={3},
  pages={581--612},
  year={2017},
  publisher={Springer}
}

@article{zeldes-etal-2025-erst,
    title = "e{RST}: A Signaled Graph Theory of Discourse Relations and Organization",
    author = "Zeldes, Amir  and
      Aoyama, Tatsuya  and
      Liu, Yang Janet  and
      Peng, Siyao  and
      Das, Debopam  and
      Gessler, Luke",
    journal = "Computational Linguistics",
    volume = "51",
    number = "1",
    month = mar,
    year = "2025",
    address = "Cambridge, MA",
    publisher = "MIT Press",
    url = "https://aclanthology.org/2025.cl-1.3/",
    doi = "10.1162/coli_a_00538",
    pages = "23--72"
}

@inproceedings{gessler-etal-2019-discourse,
    title = "A Discourse Signal Annotation System for {RST} Trees",
    author = "Gessler, Luke  and
      Liu, Yang  and
      Zeldes, Amir",
    editor = "Zeldes, Amir  and
      Das, Debopam  and
      Galani, Erick Maziero  and
      Antonio, Juliano Desiderato  and
      Iruskieta, Mikel",
    booktitle = "Proceedings of the Workshop on Discourse Relation Parsing and Treebanking 2019",
    month = jun,
    year = "2019",
    address = "Minneapolis, MN",
    publisher = "Association for Computational Linguistics",
    url = "https://aclanthology.org/W19-2708/",
    doi = "10.18653/v1/W19-2708",
    pages = "56--61"
}

@inproceedings{ju-etal-2025-dedisco,
    title = "{D}e{D}is{C}o at the {DISRPT} 2025 Shared Task: A System for Discourse Relation Classification",
    author = "Ju, Zhuoxuan  and
      Wu, Jingni  and
      Purushothama, Abhishek  and
      Zeldes, Amir",
    editor = "Braud, Chlo{\'e}  and
      Liu, Yang Janet  and
      Muller, Philippe  and
      Zeldes, Amir  and
      Li, Chuyuan",
    booktitle = "Proceedings of the 4th Shared Task on Discourse Relation Parsing and Treebanking (DISRPT 2025)",
    month = nov,
    year = "2025",
    address = "Suzhou, China",
    publisher = "Association for Computational Linguistics",
    url = "https://aclanthology.org/2025.disrpt-1.4/",
    doi = "10.18653/v1/2025.disrpt-1.4",
    pages = "48--62",
    ISBN = "979-8-89176-344-9"
}

@inproceedings{peng-etal-2022-gcdt,
    title = "{GCDT}: A {C}hinese {RST} Treebank for Multigenre and Multilingual Discourse Parsing",
    author = "Peng, Siyao  and
      Liu, Yang Janet  and
      Zeldes, Amir",
    editor = "He, Yulan  and
      Ji, Heng  and
      Li, Sujian  and
      Liu, Yang  and
      Chang, Chua-Hui",
    booktitle = "Proceedings of the 2nd Conference of the Asia-Pacific Chapter of the Association for Computational Linguistics and the 12th International Joint Conference on Natural Language Processing (Volume 2: Short Papers)",
    month = nov,
    year = "2022",
    address = "Online only",
    publisher = "Association for Computational Linguistics",
    url = "https://aclanthology.org/2022.aacl-short.47/",
    doi = "10.18653/v1/2022.aacl-short.47",
    pages = "382--391"
}

@inproceedings{CRPC-DB-Portuguese,
author = {Mendes, Am\'{a}lia and Lejeune, Pierre},
title = {{CRPC-DB} a Discourse Bank for {P}ortuguese},
year = {2022},
isbn = {978-3-030-98304-8},
publisher = {Springer-Verlag},
address = {Berlin, Heidelberg},
url = {https://doi.org/10.1007/978-3-030-98305-5_8},
doi = {10.1007/978-3-030-98305-5_8},
booktitle = {Computational Processing of the Portuguese Language: 15th International Conference, PROPOR 2022, Fortaleza, Brazil, March 21–23, 2022, Proceedings},
pages = {79–89},
numpages = {11},
location = {Fortaleza, Brazil}
}

@inproceedings{behzad-zeldes-2020-cross,
    title = "A Cross-Genre Ensemble Approach to Robust {R}eddit Part of Speech Tagging",
    author = "Behzad, Shabnam  and
      Zeldes, Amir",
    editor = {Barbaresi, Adrien  and
      Bildhauer, Felix  and
      Sch{\"a}fer, Roland  and
      Stemle, Egon},
    booktitle = "Proceedings of the 12th Web as Corpus Workshop",
    month = may,
    year = "2020",
    address = "Marseille, France",
    publisher = "European Language Resources Association",
    url = "https://aclanthology.org/2020.wac-1.7/",
    pages = "50--56",
    language = "eng",
    ISBN = "979-10-95546-68-9"
}

@inproceedings{aoyama-etal-2023-gentle,
    title = "{GENTLE}: A Genre-Diverse Multilayer Challenge Set for {E}nglish {NLP} and Linguistic Evaluation",
    author = "Aoyama, Tatsuya  and
      Behzad, Shabnam  and
      Gessler, Luke  and
      Levine, Lauren  and
      Lin, Jessica  and
      Liu, Yang Janet  and
      Peng, Siyao  and
      Zhu, Yilun  and
      Zeldes, Amir",
    editor = "Prange, Jakob  and
      Friedrich, Annemarie",
    booktitle = "Proceedings of the 17th Linguistic Annotation Workshop (LAW-XVII)",
    month = jul,
    year = "2023",
    address = "Toronto, Canada",
    publisher = "Association for Computational Linguistics",
    url = "https://aclanthology.org/2023.law-1.17/",
    doi = "10.18653/v1/2023.law-1.17",
    pages = "166--178"
}

@inproceedings{liu-etal-2024-gdtb,
    title = "{GDTB}: Genre Diverse Data for {E}nglish Shallow Discourse Parsing across Modalities, Text Types, and Domains",
    author = "Liu, Yang Janet  and
      Aoyama, Tatsuya  and
      Scivetti, Wesley  and
      Zhu, Yilun  and
      Behzad, Shabnam  and
      Levine, Lauren Elizabeth  and
      Lin, Jessica  and
      Tiwari, Devika  and
      Zeldes, Amir",
    editor = "Al-Onaizan, Yaser  and
      Bansal, Mohit  and
      Chen, Yun-Nung",
    booktitle = "Proceedings of the 2024 Conference on Empirical Methods in Natural Language Processing",
    month = nov,
    year = "2024",
    address = "Miami, Florida, USA",
    publisher = "Association for Computational Linguistics",
    url = "https://aclanthology.org/2024.emnlp-main.684/",
    doi = "10.18653/v1/2024.emnlp-main.684",
    pages = "12287--12303"
}

@inproceedings{bauer-etal-2023-semgrex,
    title = "Semgrex and Ssurgeon, Searching and Manipulating Dependency Graphs",
    author = "Bauer, John  and
      Kiddon, Chlo{\'e}  and
      Yeh, Eric  and
      Shan, Alex  and
      D. Manning, Christopher",
    editor = {Dakota, Daniel  and
      Evang, Kilian  and
      K{\"u}bler, Sandra  and
      Levin, Lori},
    booktitle = "Proceedings of the 21st International Workshop on Treebanks and Linguistic Theories (TLT, GURT/SyntaxFest 2023)",
    month = mar,
    year = "2023",
    address = "Washington, D.C.",
    publisher = "Association for Computational Linguistics",
    url = "https://aclanthology.org/2023.tlt-1.7/",
    pages = "67--73"
}

@inproceedings{tamburini-2017-semgrex,
    title = "Semgrex-Plus: a Tool for Automatic Dependency-Graph Rewriting",
    author = "Tamburini, Fabio",
    editor = "Montemagni, Simonetta  and
      Nivre, Joakim",
    booktitle = "Proceedings of the Fourth International Conference on Dependency Linguistics (Depling 2017)",
    month = sep,
    year = "2017",
    address = "Pisa, Italy",
    publisher = {Link{\"o}ping University Electronic Press},
    url = "https://aclanthology.org/W17-6528/",
    pages = "248--254"
}

@inproceedings{german:rst:pcc,
    title = "{P}otsdam Commentary Corpus 2.0: Annotation for Discourse Research",
    author = "Stede, Manfred  and
      Neumann, Arne",
    editor = "Calzolari, Nicoletta  and
      Choukri, Khalid  and
      Declerck, Thierry  and
      Loftsson, Hrafn  and
      Maegaard, Bente  and
      Mariani, Joseph  and
      Moreno, Asuncion  and
      Odijk, Jan  and
      Piperidis, Stelios",
    booktitle = "Proceedings of the Ninth International Conference on Language Resources and Evaluation ({LREC}'14)",
    month = may,
    year = "2014",
    address = "Reykjavik, Iceland",
    publisher = "European Language Resources Association (ELRA)",
    url = "https://aclanthology.org/L14-1468/",
    pages = "925--929"
}

@article{english:rst:covdtb,
    title = "Out-of-Domain Discourse Dependency Parsing via Bootstrapping: An Empirical Analysis on Its Effectiveness and Limitation",
    author = "Nishida, Noriki  and
      Matsumoto, Yuji",
    editor = "Roark, Brian  and
      Nenkova, Ani",
    journal = "Transactions of the Association for Computational Linguistics",
    volume = "10",
    year = "2022",
    address = "Cambridge, MA",
    publisher = "MIT Press",
    url = "https://aclanthology.org/2022.tacl-1.8/",
    doi = "10.1162/tacl_a_00451",
    pages = "127--144"
}

@inproceedings{english:dep:scidtb,
    title = "{S}ci{DTB}: Discourse Dependency {T}ree{B}ank for Scientific Abstracts",
    author = "Yang, An  and
      Li, Sujian",
    editor = "Gurevych, Iryna  and
      Miyao, Yusuke",
    booktitle = "Proceedings of the 56th Annual Meeting of the Association for Computational Linguistics (Volume 2: Short Papers)",
    month = jul,
    year = "2018",
    address = "Melbourne, Australia",
    publisher = "Association for Computational Linguistics",
    url = "https://aclanthology.org/P18-2071/",
    doi = "10.18653/v1/P18-2071",
    pages = "444--449"
}

@InCollection{english:rst:rstdt,
  title                    = {Building a Discourse-Tagged Corpus in the Framework of Rhetorical Structure Theory},
  author                   = {Lynn Carlson and Daniel Marcu and Mary Ellen Okurowski},
  booktitle                = {Current and New Directions in Discourse and Dialogue},
  publisher                = {Kluwer},
  year                     = {2003},

  address                  = {Dordrecht},
  pages                    = {85-112},
  series                   = {Text, Speech and Language Technology 22}
}

@inproceedings{english:sdrt:stac,
    title = "Discourse Structure and Dialogue Acts in Multiparty Dialogue: the {STAC} Corpus",
    author = "Asher, Nicholas  and
      Hunter, Julie  and
      Morey, Mathieu  and
      Farah, Benamara  and
      Afantenos, Stergos",
    editor = "Calzolari, Nicoletta  and
      Choukri, Khalid  and
      Declerck, Thierry  and
      Goggi, Sara  and
      Grobelnik, Marko  and
      Maegaard, Bente  and
      Mariani, Joseph  and
      Mazo, Helene  and
      Moreno, Asuncion  and
      Odijk, Jan  and
      Piperidis, Stelios",
    booktitle = "Proceedings of the Tenth International Conference on Language Resources and Evaluation ({LREC}'16)",
    month = may,
    year = "2016",
    address = "Portoro{\v{z}}, Slovenia",
    publisher = "European Language Resources Association (ELRA)",
    url = "https://aclanthology.org/L16-1432/",
    pages = "2721--2727"
}

@misc{basque:rst:ert,
  author = "Mikel Iruskieta and Mar\'{i}a Jes\'{u}s Aranzabe and Arantza Diaz de Ilarraza and Itziar Gonzalez-Dios and Mikel Lersundi and Oier Lopez de Lacalle",
  title = "The {RST} {B}asque {TreeBank}",
  year = 2012,
  publisher = "self",
  language = "basque", 
  url = "http://ixa2.si.ehu.eus/diskurtsoa/en/",
}

@misc{persian:rst:prstc,
  author       = {Sara Shahmohammadi and Hadi Veisi and Ali Darzi},
  title        = {Persian Rhetorical Structure Theory},
  year         = {2021},
  month        = jun,
  doi          = {10.48550/arXiv.2106.13833},
  eprint       = {2106.13833},
  archivePrefix = {arXiv},
  primaryClass = {cs.CL},
  url          = {https://arxiv.org/abs/2106.13833}
}

@inproceedings{french:sdrt:annodis,
    title = "An empirical resource for discovering cognitive principles of discourse organisation: {T}he {ANNODIS} corpus",
    author = "Afantenos, Stergos  and
      Asher, Nicholas  and
      Benamara, Farah  and
      Bras, Myriam  and
      Fabre, C{\'e}cile  and
      Ho-dac, Mai  and
      Draoulec, Anne Le  and
      Muller, Philippe  and
      P{\'e}ry-Woodley, Marie-Paule  and
      Pr{\'e}vot, Laurent  and
      Rebeyrolles, Josette  and
      Tanguy, Ludovic  and
      Vergez-Couret, Marianne  and
      Vieu, Laure",
    editor = "Calzolari, Nicoletta  and
      Choukri, Khalid  and
      Declerck, Thierry  and
      Do{\u{g}}an, Mehmet U{\u{g}}ur  and
      Maegaard, Bente  and
      Mariani, Joseph  and
      Moreno, Asuncion  and
      Odijk, Jan  and
      Piperidis, Stelios",
    booktitle = "Proceedings of the Eighth International Conference on Language Resources and Evaluation ({LREC}'12)",
    month = may,
    year = "2012",
    address = "Istanbul, Turkey",
    publisher = "European Language Resources Association (ELRA)",
    url = "https://aclanthology.org/L12-1498/",
    pages = "2727--2734"
}

@inproceedings{dutch:rst:nldt,
    title = "Multi-Layer Discourse Annotation of a {D}utch Text Corpus",
    author = "Redeker, Gisela  and
      Berzl{\'a}novich, Ildik{\'o}  and
      van der Vliet, Nynke  and
      Bouma, Gosse  and
      Egg, Markus",
    editor = "Calzolari, Nicoletta  and
      Choukri, Khalid  and
      Declerck, Thierry  B.
      Do{\u{g}}an, Mehmet U{\u{g}}ur  and
      Maegaard, Bente  and
      Mariani, Joseph  and
      Moreno, Asuncion  and
      Odijk, Jan  and
      Piperidis, Stelios",
    booktitle = "Proceedings of the Eighth International Conference on Language Resources and Evaluation ({LREC}'12)",
    month = may,
    year = "2012",
    address = "Istanbul, Turkey",
    publisher = "European Language Resources Association (ELRA)",
    url = "https://aclanthology.org/L12-1528/",
    pages = "2820--2825"
}

@inproceedings{portuguese:rst:cstn,
  author    = {Paula C. F. Cardoso and Erick G. Maziero and Maria Luc{\'\i}a R. Castro Jorge and Eloize M. R. Seno and Ariani Di Felippo and Lucia H. M. Rino and Maria das Gra{\c c}as V. Nunes and Thiago A. S. Pardo},
  title     = {{CSTNews} -- A Discourse-Annotated Corpus for Single and Multi-Document Summarization of News Texts in Brazilian Portuguese},
  booktitle = {Anais do III Workshop ``A RST e os Estudos do Texto''},
  pages     = {88--105},
  address   = {Cuiab{\'a}, MT, Brasil},
  month     = oct,
  year      = {2011},
  publisher = {Sociedade Brasileira de Computa{\c c}{\~a}o}
}

@inproceedings{russian:rst:rrt,
  author    = {Dina Pisarevskaya and Margarita Ananyeva and Maria Kobozeva and Alexander Nasedkin and Sofia Nikiforova and Irina Pavlova and Alexey Shelepov},
  title     = {Towards Building a Discourse-Annotated Corpus of Russian},
  booktitle = {Proceedings of the 23rd International Conference on Computational Linguistics and Intellectual Technologies ``Dialogue-2017''},
  year      = {2017},
  address   = {Moscow, Russia},
  month     = jun
}

@inproceedings{spanishchinese:rst:sctb,
    title = "The {RST} {S}panish-{C}hinese Treebank",
    author = "Cao, Shuyuan  and
      da Cunha, Iria  and
      Iruskieta, Mikel",
    editor = "Savary, Agata  and
      Ramisch, Carlos  and
      Hwang, Jena D.  and
      Schneider, Nathan  and
      Andresen, Melanie  and
      Pradhan, Sameer  and
      Petruck, Miriam R. L.",
    booktitle = "Proceedings of the Joint Workshop on Linguistic Annotation, Multiword Expressions and Constructions ({LAW}-{MWE}-{C}x{G}-2018)",
    month = aug,
    year = "2018",
    address = "Santa Fe, New Mexico, USA",
    publisher = "Association for Computational Linguistics",
    url = "https://aclanthology.org/W18-4917/",
    pages = "156--166"
}

@inproceedings{chinese:dep:scidtb,
    title = "Unifying Discourse Resources with Dependency Framework",
    author = "Yi, Cheng  and
      Sujian, Li  and
      Yueyuan, Li",
    editor = "Li, Sheng  and
      Sun, Maosong  and
      Liu, Yang  and
      Wu, Hua  and
      Liu, Kang  and
      Che, Wanxiang  and
      He, Shizhu  and
      Rao, Gaoqi",
    booktitle = "Proceedings of the 20th Chinese National Conference on Computational Linguistics",
    month = aug,
    year = "2021",
    address = "Huhhot, China",
    publisher = "Chinese Information Processing Society of China",
    url = "https://aclanthology.org/2021.ccl-1.94/",
    pages = "1058--1065",
    language = "eng"
}

@article{multi:pdtb:ted,
  author    = {Deniz Zeyrek and Am{\'a}lia Mendes and Yulia Grishina and Murathan Kurfal{\i} and Samuel Gibbon and Maciej Ogrodniczuk},
  title     = {TED Multilingual Discourse Bank (TED-MDB): a parallel corpus annotated in the PDTB style},
  journal   = {Language Resources and Evaluation},
  year      = {2020},
  volume    = {54},
  number    = {2},
  pages     = {587--613},
  doi       = {10.1007/s10579-019-09445-9},
  url       = {https://doi.org/10.1007/s10579-019-09445-9},
  issn      = {1574-0218}
}

@misc{italian:pdtb:luna,
  author = "Tonelli, Sara  and
      Riccardi, Giuseppe  and
      Prasad, Rashmi  and
      Joshi, Aravind and Stepanov, Evgeny A. and Chowdhury, Shammur Absar",
  title = "{LUNA Corpus Discourse Data Set}",
  year = 2010,
  publisher = "ELRA",
  language = "italian", 
  url = "http://universal.elra.info/product\_info.php?-cPath=37\_38\&products\_id=1832",
}

@inproceedings{turkish:pdtb:tdb,
    title = "{TDB} 1.1: Extensions on {T}urkish Discourse Bank",
    author = "Zeyrek, Deniz  and
      Kurfal{\i}, Murathan",
    editor = "Schneider, Nathan  and
      Xue, Nianwen",
    booktitle = "Proceedings of the 11th Linguistic Annotation Workshop",
    month = apr,
    year = "2017",
    address = "Valencia, Spain",
    publisher = "Association for Computational Linguistics",
    url = "https://aclanthology.org/W17-0809/",
    doi = "10.18653/v1/W17-0809",
    pages = "76--81"
}

@misc{chinese:pdtb:cdtb,
  author = "Yuping Zhou and Jill Lu and Jennifer Zhang and Nianwen Xue",
  title = "{Chinese {D}iscourse {T}reebank 0.5}",
  year = 2014,
  publisher = "LDC",
  language = "chinese", 
  url = "https://catalog.ldc.upenn.edu/LDC2014T21",
}

@article{potter2008interactional,
  author  = {Andrew Potter},
  title   = {Interactional coherence in asynchronous learning networks: A rhetorical approach},
  journal = {The Internet and Higher Education},
  volume  = {11},
  number  = {2},
  pages   = {87--97},
  year    = {2008}
}

@inproceedings{zaczynska-stede-2024-rhetorical,
    title = "Rhetorical Strategies in the {UN} Security Council: {R}hetorical {S}tructure {T}heory and Conflicts",
    author = "Zaczynska, Karolina  and
      Stede, Manfred",
    editor = "Kawahara, Tatsuya  and
      Demberg, Vera  and
      Ultes, Stefan  and
      Inoue, Koji  and
      Mehri, Shikib  and
      Howcroft, David  and
      Komatani, Kazunori",
    booktitle = "Proceedings of the 25th Annual Meeting of the Special Interest Group on Discourse and Dialogue",
    month = sep,
    year = "2024",
    address = "Kyoto, Japan",
    publisher = "Association for Computational Linguistics",
    url = "https://aclanthology.org/2024.sigdial-1.2/",
    doi = "10.18653/v1/2024.sigdial-1.2",
    pages = "15--28"
}

@inproceedings{thompson-etal-2024-discourse,
    title = "Discourse Structure for the {M}inecraft Corpus",
    author = "Thompson, Kate  and
      Hunter, Julie  and
      Asher, Nicholas",
    editor = "Calzolari, Nicoletta  and
      Kan, Min-Yen  and
      Hoste, Veronique  and
      Lenci, Alessandro  and
      Sakti, Sakriani  and
      Xue, Nianwen",
    booktitle = "Proceedings of the 2024 Joint International Conference on Computational Linguistics, Language Resources and Evaluation (LREC-COLING 2024)",
    month = may,
    year = "2024",
    address = "Torino, Italia",
    publisher = "ELRA and ICCL",
    url = "https://aclanthology.org/2024.lrec-main.444/",
    pages = "4957--4967"
}

@article{scholman2025disconaija,
  title={{DiscoNaija}: {A} discourse-annotated parallel {Nigerian Pidgin-English} corpus},
  author={Scholman, Merel CJ and Marchal, Marian and Brown, AriaRay and Demberg, Vera},
  journal={Language Resources and Evaluation},
  pages={1--37},
  year={2025},
  publisher={Springer}
}

@inproceedings{ogrodniczuk-etal-2024-polish-discourse,
    title = "{P}olish Discourse Corpus ({PDC}): Corpus Design, {ISO}-Compliant Annotation, Data Highlights, and Parser Development",
    author = "Ogrodniczuk, Maciej  and
      Tomaszewska, Aleksandra  and
      Ziembicki, Daniel  and
      {\.Z}urowski, Sebastian  and
      Tuora, Ryszard  and
      Zwierzchowska, Aleksandra",
    editor = "Calzolari, Nicoletta  and
      Kan, Min-Yen  and
      Hoste, Veronique  and
      Lenci, Alessandro  and
      Sakti, Sakriani  and
      Xue, Nianwen",
    booktitle = "Proceedings of the 2024 Joint International Conference on Computational Linguistics, Language Resources and Evaluation (LREC-COLING 2024)",
    month = may,
    year = "2024",
    address = "Torino, Italia",
    publisher = "ELRA and ICCL",
    url = "https://aclanthology.org/2024.lrec-main.1123/",
    pages = "12829--12835"
}

@article{PrasertsomEtAl2024,
    author = {Prasertsom, Ponrawee and Jaroonpol, Apiwat and Rutherford, Attapol T.},
    title = {The {Thai Discourse Treebank}: {A}nnotating and Classifying {T}hai Discourse Connectives},
    journal = {Transactions of the Association for Computational Linguistics},
    volume = {12},
    pages = {613--629},
    year = {2024},
    month = {05},
    issn = {2307-387X},
    doi = {10.1162/tacl_a_00650},
    url = {https://doi.org/10.1162/tacl_a_00650},
    eprint = {https://direct.mit.edu/tacl/article-pdf/doi/10.1162/tacl_a_00650/2369332/tacl_a_00650.pdf},
}

\appendix

\section{Dataset details}
\label{sec:appendix-datasets}

Table \ref{tab:datasets} gives an overview of the datasets that are currently searchable using the system. Datasets marked by an asterisk ($\ast$) require LDC licenses; annotations for this data and a small subset of \texttt{eng.erst.gum} coming from Reddit \cite{behzad-zeldes-2020-cross} can be obtained from the DISRPT shared task repository, along with scripts to reconstruct the underlying text.

\begin{table*}[tbh]
\resizebox{\textwidth}{!}{
\begin{tabular}{lllllllll}
\toprule
\textbf{Corpus}     & \textbf{Language} & \textbf{Framework} & \textbf{Labels} & \textbf{Relations} & \textbf{Sentences} & \textbf{Tokens} & \textbf{Documents} & \textbf{Signals} \\
\midrule
ces.rst.crdt        & Czech         & RST                & 17              & 1,249              & 835                & 14,664          & 54                 & --               \\
deu.pdtb.pcc        & German        & PDTB               & 11              & 2,109              & 2,193              & 33,222          & 176                & types            \\
deu.rst.pcc         & German        & RST                & 16              & 2,882              & 1,944              & 32,836          & 176                & --               \\
eng.dep.covdtb      & English       & dependencies       & 11              & 4,985              & 2,343              & 60,907          & 300                & --               \\
eng.dep.scidtb      & English       & dependencies       & 14              & 9,903              & 4,202              & 102,534         & 798                & --               \\
eng.erst.gentle     & English       & eRST               & 17              & 2,552              & 1,334              & 17,979          & 26                 & subtypes         \\
eng.erst.gum        & English       & eRST               & 17              & 30,747             & 14,158             & 254,890         & 255                & subtypes         \\
eng.pdtb.gentle     & English       & PDTB               & 12              & 786                & 1,334              & 17,979          & 26                 & types            \\
eng.pdtb.gum        & English       & PDTB               & 13              & 13,879             & 14,158             & 254,890         & 255                & types            \\
$\ast$eng.pdtb.pdtb       & English       & PDTB               & 13              & 47,792             & 48,630             & 1,173,379       & 2,162              & types            \\
eng.pdtb.tedm       & English       & PDTB               & 13              & 529                & 381                & 8,185           & 6                  & types            \\
eng.rst.oll         & English       & RST                & 17              & 2,751              & 2,156              & 46,471          & 327                & --               \\
$\ast$eng.rst.rstdt       & English       & RST                & 17              & 19,778             & 8,318              & 208,912         & 385                & --               \\
eng.rst.sts         & English       & RST                & 17              & 3,058              & 2,591              & 71,206          & 150                & --               \\
eng.rst.umuc        & English       & RST                & 15              & 4,997              & 2,424              & 61,590          & 87                 & --               \\
eng.sdrt.msdc       & English       & SDRT               & 10              & 27,848             & 14,744             & 231,352         & 440                & --               \\
eng.sdrt.stac       & English       & SDRT               & 11              & 12,271             & 7,394              & 52,271          & 1,101              & --               \\
eus.rst.ert         & Basque        & RST                & 16              & 3,632              & 2,380              & 45,780          & 164                & --               \\
fas.rst.prstc       & Farsi         & RST                & 14              & 5,191              & 2,179              & 66,926          & 150                & --               \\
fra.sdrt.annodis    & French        & SDRT               & 12              & 3,321              & 1,507              & 32,699          & 86                 & --               \\
ita.pdtb.luna       & Italian       & PDTB               & 11              & 1,525              & 3,750              & 25,242          & 60                 & types            \\
nld.rst.nldt        & Dutch         & RST                & 16              & 2,264              & 1,651              & 24,898          & 80                 & --               \\
pcm.pdtb.disconaija & Naija         & PDTB               & 13              & 9,903              & 9,242              & 140,729         & 176                & types            \\
pol.iso.pdc         & Polish        & ISO                & 12              & 8,543              & 9,142              & 156,980         & 556                & types            \\
por.pdtb.crpc       & Portuguese    & PDTB               & 12              & 11,327             & 5,194              & 186,849         & 302                & types            \\
por.pdtb.tedm       & Portuguese    & PDTB               & 13              & 554                & 394                & 8,190           & 6                  & types            \\
por.rst.cstn        & Portuguese    & RST                & 15              & 4,993              & 2,221              & 63,332          & 140                & --               \\
rus.rst.rrt         & Russian       & RST                & 15              & 25,095             & 13,131             & 262,495         & 234                & --               \\
spa.rst.rststb      & Spanish       & RST                & 16              & 3,049              & 2,089              & 58,717          & 267                & --               \\
spa.rst.sctb        & Spanish       & RST                & 16              & 692                & 516                & 16,515          & 50                 & --               \\
tha.pdtb.tdtb       & Thai          & PDTB               & 12              & 10,861             & 6,534              & 256,523         & 180                & --               \\
$\ast$tur.pdtb.tdb        & Turkish       & PDTB               & 13              & 3,176              & 31,197             & 496,358         & 197                & --               \\
tur.pdtb.tedm       & Turkish       & PDTB               & 13              & 574                & 410                & 6,286           & 6                  & types            \\
zho.dep.scidtb      & Mandarin      & dependencies       & 14              & 1,297              & 500                & 18,761          & 109                & --               \\
zho.pdtb.cdtb       & Mandarin      & PDTB               & 9               & 5,270              & 2,891              & 73,314          & 164                & --               \\
zho.pdtb.ted        & Mandarin      & PDTB               & 15              & 13,308             & 8,671              & 181,910         & 72                 & types            \\
zho.rst.gcdt        & Mandarin      & RST                & 17              & 8,413              & 2,692              & 62,905          & 50                 & --               \\
zho.rst.sctb        & Mandarin      & RST                & 17              & 692                & 580                & 15,496          & 50                 & --               \\
\midrule
\textbf{Total}               & 16            & 6                  & 17              & 311,796            & 257,705            & 5,139,564       & 9,890              & 14 datasets   \\  
\bottomrule
\end{tabular}
}
\caption{DISRPT 2025 datasets searchable in DiscoExplorer ($\ast$ marks datasets requiring an LDC license).}
\label{tab:datasets}
\end{table*}

\begin{figure*}[tbh]
\centering
\includegraphics[width=\textwidth, frame]{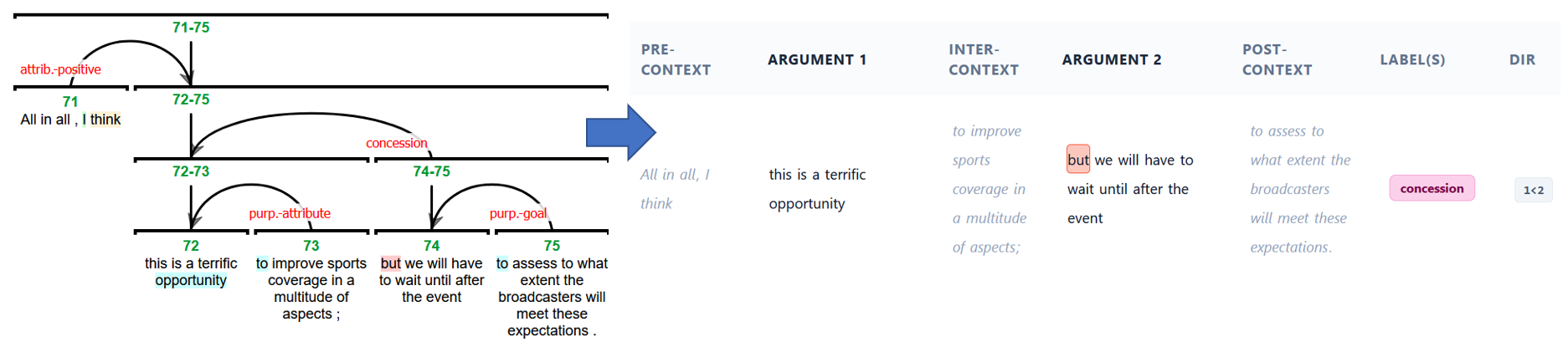}
\caption{Original eRST graph fragment for a \textsc{concession} relation, visualized using rstWeb \cite{gessler-etal-2019-discourse} and the corresponding output in DiscoExplorer.}
\label{fig:data-conversion}
\end{figure*}

Additional datasets can be added to the system as long as they conform to the DISRPT shared task format, meaning that relations are serialized in the \texttt{.rels} format and token annotations are available in a corresponding \texttt{.conllu} file. In particular, the DISRPT format assumes that relations apply between flat spans of text, meaning that hierarchical information as found in formalisms such as RST is lost. Instead, relations are interpreted as a dependency conversion of constituent structures, as illustrated in Figure \ref{fig:data-conversion}. 

The figure shows the system representation of a single relation, in this case a \textsc{concession} between two head units: 

\ex. \a. $[$\textit{this is a terrific opportunity}$]$
\b. $[$\textit{\textbf{but} we will have to wait until after the event}$]$ 

These units are only parts of the sentences they come from, as shown in the eRST  graph fragment. Modifiers of those units which appear in the same sentences are represented in DiscoExplorer as pre-, inter- and post-context, depending on whether they appear before the first argument, between the two arguments, or after the second. 

Note also that while the eRST graph on the left highlights multiple signals, only the red discourse marker ``but'' is attached to the \textsc{concession} relation, and that same signal is visualized in the DiscoExplorer search results. The cyan highlighted syntactic signals for \textsc{purpose} relations (`opportunity .. to improve' and `wait .. to assess') belong to those respective relations, and would be highlighted in queries actually retrieving the associated relation, rather than being highlighted in a query retrieving a different relation that happens to overlap the same text.

\section{Additional use cases}\label{sec:additional}

\begin{figure}[h!tb]
\centering
\includegraphics[width=\columnwidth, frame]{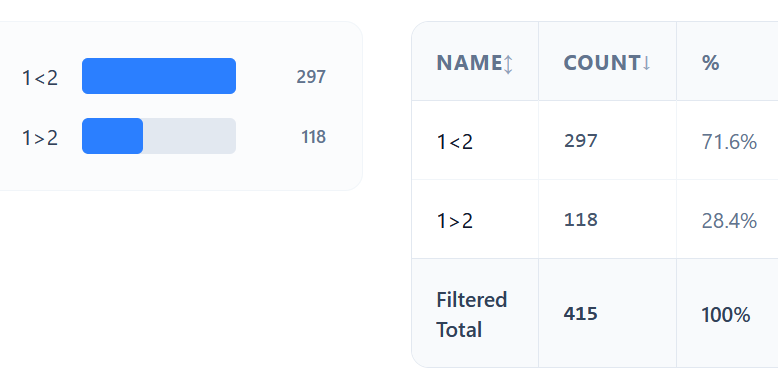}
\caption{Frequencies of \textsc{temporal} `when' clauses in left-to-right vs. right-to-left directions}
\label{fig:when}
\end{figure}

\begin{figure*}[t!]
\centering
\begin{subfigure}[t!]{0.91\textwidth}
\centering
\includegraphics[width=0.9\textwidth, frame]{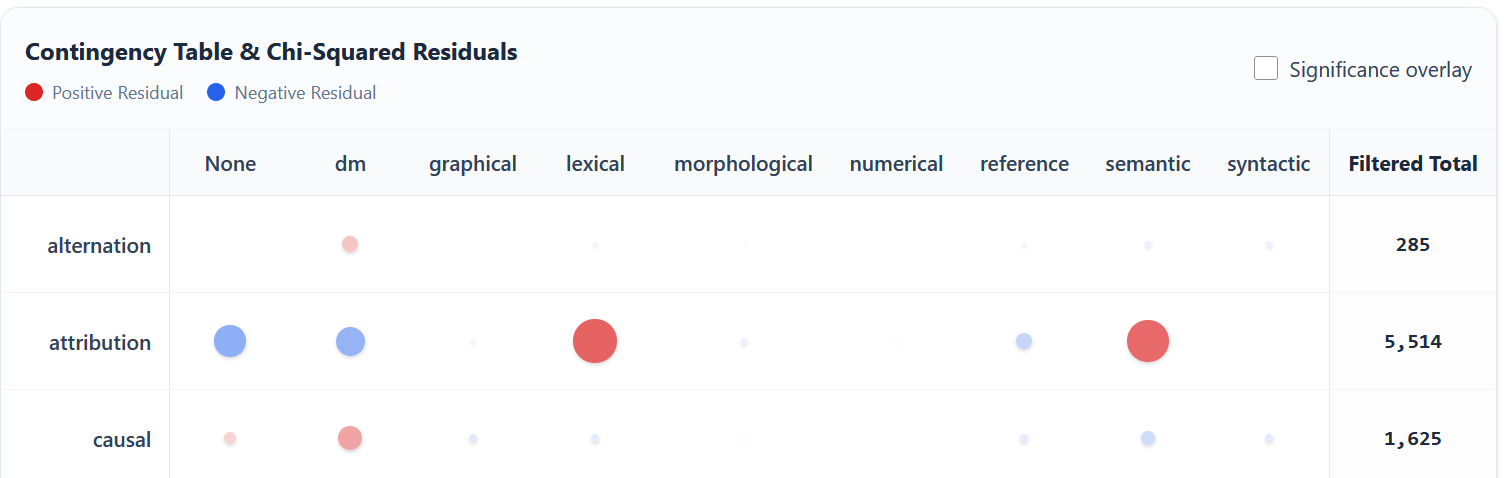}
\caption{Residual plot for DISRPT labels versus major signal types.} \label{fig:sigtypes}
\end{subfigure}

\begin{subfigure}[t!]{0.9\textwidth}
\centering
\includegraphics[width=0.9\textwidth, frame]{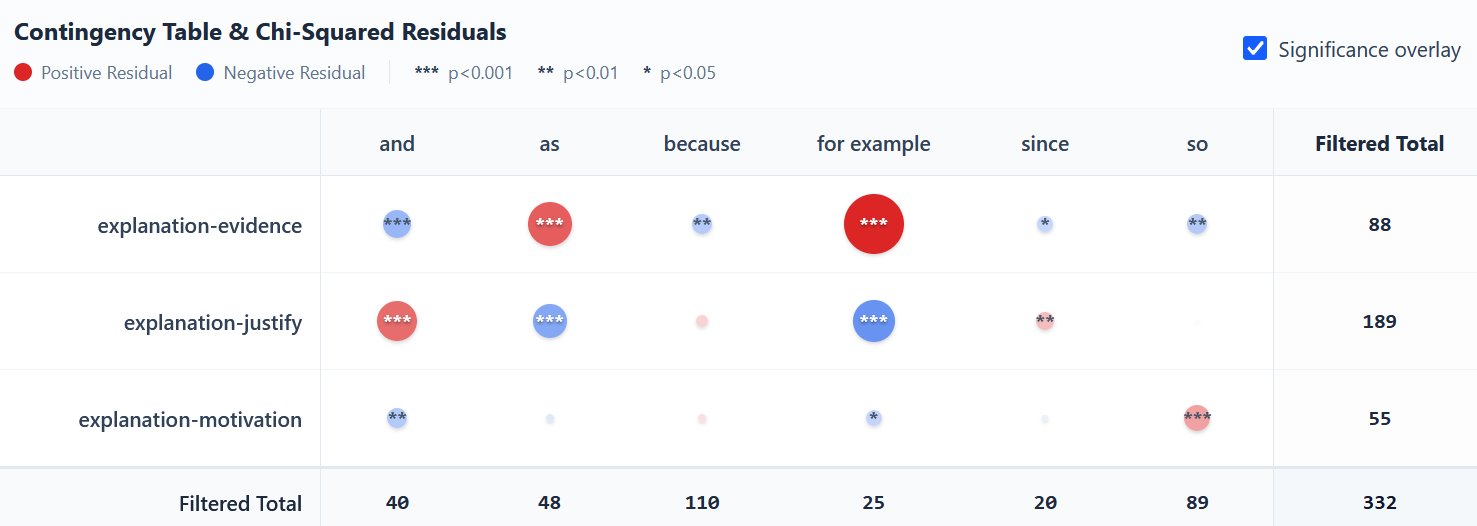}
\caption{Residual plot for discourse markers appearing 20+ times versus \textsc{explanation} relation labels} \label{fig:dms}
\end{subfigure}

 \caption{Contingency tables for signal types and subtypes in \texttt{eng.erst.gum}.}\label{fig:panels}

\end{figure*}

In addition to exploring relation labels, DiscoExplorer can be used to study the distribution of relation signals (in datasets with signal annotations, indicated in the dataset list by a lightning bolt icon in Figure \ref{fig:search-interface}), relation directions and signal subtypes. Figure \ref{fig:sigtypes} shows a breakdown of relation 

Figure \ref{fig:when} shows the disparity of right-to-left versus left-to-right temporal relations signaled by `when' in English, where source of the relation tends to precede the target, but not always -- the proportion is about 7:3 in favor of placing the \textsc{temporal} clause first.

Figure \ref{fig:panels} shows two panels demonstrating the breakdown of signaling devices. Panel \ref{fig:sigtypes} shows major signal types signaling three relations classes: \textsc{alternation} relations (such as `A or B') are typically signaled by a discourse marker (dm) such as `(either) or', `(or) else', `alternatively' etc.; \textsc{attribution} relations, indicating the source of information, are typically signaled semantically by the presence of a speaker noun phrase, or lexically through a speech verb, but very rarely using a dm (for example `\textbf{As} Smith had it, ...') or with no signals (implicitly). By contrast \textsc{causal} relations usually appear with a dm, and appear with no explicit signaling more often than the other relations (column `None').

Finally, Panel \ref{fig:dms} shows discourse marker subtypes used to signal three original labels for \textsc{explanation} relations in the data, using a cutoff to show only items appearing 20 or more times. Although the DISRPT relation labels do not include subtypes, we can get breakdowns for relation subtypes if the original labels of the underlying dataset include them, which is the case here. The biggest disparity is the preference of \textsc{explanation-evidence} relations to be marked by `for example' and `as'. By contrast, \textsc{explanation-justify} favors the use of `and', as in \ref{ex:and-justify}.

\ex. $[$\textit{The record is replete with case law that says exactly that,}$]_1$ $[$\textbf{\textit{and}} \textit{I'm not here to dispute that today.}$]_2$\label{ex:and-justify}

Meanwhile \textsc{explanation-motivation} relations appear disproportionally often with `so', attempting to convince someone to do something using a supporting argument, as in \ref{ex:so}.

\ex. $[$\textit{Good jokes have a lot of details and personality,}$]_1$ $[$\textbf{\textit{so}} \textit{don’t be afraid to embellish.}$]_2$\label{ex:so}

However using the interface, it is easy to find examples of `so' as a discourse marker with any of the three labels.

\section{Resource consumption}\label{sec:performance}

While the interface runs very quickly and requires no server-side compute resources to run, an anonymous reviewer has inquired about the memory footprint of loading a large dataset. This is difficult to quantify exactly, since we cannot trivially access browser memory management internals, but to get a rough benchmark, we ran Chrome on a Windows 11 64 bit machine and compared memory usage for the browser with different datasets loaded, as shown in Table \ref{tab:memory}.

\begin{table}[h!tb]
\resizebox{\columnwidth}{!}{%
\begin{tabular}{lrrrr}
\toprule
\textbf{scenario}             & \textbf{tokens}        & \textbf{relations} & \textbf{memory} & \textbf{$\Delta$ idle} \\
\midrule
\textit{Chrome idle}          & 0                      & 0                  & 305.9 MB        & 0                   \\
\texttt{eng.erst.gum} \textit{loaded}           & 273,257                & 33,390             & 709.2 MB        & +403.3              \\
\texttt{eng.pdtb.pdtb} \textit{loaded} & 1,173,379              & 47,792             & 750.1 MB        & +444.2              \\
\textit{compare}              & \multicolumn{1}{c}{--} & 81,182             & 770.4 MB        & +464.5             \\
\bottomrule

\end{tabular}
}
\caption{Memory usage in several scenarios using Chrome.}
\label{tab:memory}
\end{table}

The dataset `loaded' state by default means that a query runs to retrieve all relations, since no filter has been selected. As the table shows, loading a fairly large and richly annotated corpus such as \texttt{eng.erst.gum} requires about 400 MB of RAM. The largest dataset in tokens and relations, \texttt{eng.pdtb.pdtb}, takes only slightly more, at 444 MB -- we suspect the small difference is due to the amount of space taken up by the signal annotations in the former dataset, which the latter lacks. Using the comparison function on the two datasets does not add much more memory ($\sim$465 MB total). We suspect that this is because a new full result set is not actually loaded into memory - only the information being compared, by default the relation label statistics, is added to the main memory, while indexing specific matching tokens or sentence spans is unnecessary for the comparison data, since no detailed search results are shown for the second dataset.

\section{Corpus resources}

The system described in this paper would be useless without the datasets it makes searchable. In addition to the datasets and papers cited above, we would like to acknowledge the projects that have produced the remaining datasets in Table \ref{tab:datasets}, all of which can be searched in our publicly available instance of DiscoExplorer and are available under their original licenses from the DISRPT shared task:

\begin{itemize}
    \item \texttt{deu.rst.pcc} and \texttt{deu.pdtb.pcc} -- the Potsdam Commentary Corpus, \citet{german:rst:pcc}
    \item \texttt{eng.dep.covdtb} -- the COVID-29 Discourse Treebank, \citet{english:rst:covdtb}
    \item \texttt{eng.dep.scidtb} -- SciDTB, \citet{english:dep:scidtb}
    \item \texttt{eng.pdtb.tedm}, \texttt{por.pdtb.tedm} and \texttt{tur.pdtb.tedm} -- TED Multilingual Discourse Bank, \citet{multi:pdtb:ted}
    \item \texttt{end.rst.oll} and \texttt{end.rst.sts} -- RST Online-Learning and Science, Technology, and Society corpora, \citet{potter2008interactional}
    \item \texttt{end.rst.rstdt} -- RST Discourse Treebank, \citet{english:rst:rstdt}
    \item \texttt{end.rst.umuc} -- University of Potsdam Multilayer UNSC Corpus, \citet{zaczynska-stede-2024-rhetorical}
    \item \texttt{eng.sdrt.msdc} -- Minecraft Structured Dialogue Corpus, \citet{thompson-etal-2024-discourse}
    \item \texttt{eng.sdrt.stac} -- Strategic Conversation Corpus, \citet{english:sdrt:stac}
    \item \texttt{eus.rst.ert} -- Basque RST Treebank, \citet{basque:rst:ert}
    \item \texttt{fas.rst.prstc} -- Persian RST Corpus, \citet{persian:rst:prstc}
    \item \texttt{fra.sdrt.annodis} -- the ANNODIS corpus, \citet{french:sdrt:annodis}
    \item \texttt{ita.pdtb.luna} -- LUNA corpus, \citet{italian:pdtb:luna}
    \item \texttt{nld.rst.nldt} -- Dutch DTB, \citet{dutch:rst:nldt}
    \item \texttt{pcm.pdtb.disconaija} -- DiscoNaija corpus, \citet{scholman2025disconaija}    
    \item \texttt{pol.iso.pdc} -- Polish Discourse Corpus, \citet{ogrodniczuk-etal-2024-polish-discourse}    
    \item \texttt{por.rst.cstn} -- CST News Corpus, \cite{portuguese:rst:cstn}
    \item \texttt{rus.rst.rrt} -- Russian RST Treebank, \citet{russian:rst:rrt}
    \item \texttt{tha.pdtb.tdtb} -- Thai Discourse Treebank, \citet{PrasertsomEtAl2024}
    \item \texttt{tur.pdtb.tdb} -- Turkish Discourse Bank, \citet{turkish:pdtb:tdb}
    \item \texttt{spa.rst.sctb} and \texttt{zho.rst.sctb} -- the RST Spanish-Chinese Treebank, \citet{spanishchinese:rst:sctb}
    \item \texttt{zho.pdtb.cdtb} -- Chinese Discourse Treebank, \citet{chinese:pdtb:cdtb}
    \item \texttt{zho.dep.scidtb} -- Chinese SciDTB, \citet{chinese:dep:scidtb}
    
\end{itemize}

\end{document}